%% file: main.tex
\documentclass{article}

\usepackage{microtype}
\usepackage{graphicx}
\usepackage{subfigure}
\usepackage{booktabs} 
\usepackage{multirow}

\usepackage{hyperref}

\usepackage{amssymb}
\usepackage{pifont}
\newcommand{\cmark}{\ding{51}}%
\newcommand{\xmark}{\scalebox{0.7}{\ding{55}}}%

\newcommand\textcite[1]{\citeauthor{#1} \yrcite{#1}}
\usepackage{fancyhdr,graphicx,amsmath,amssymb}
\usepackage[ruled,vlined]{algorithm2e}
\include{pythonlisting}

\usepackage[accepted]{icml2020}

\usepackage[font=small,skip=3pt]{caption}
\usepackage[shortlabels,inline]{enumitem}

\icmltitlerunning{Defining Benchmarks for Continual Few-Shot Learning}

\definecolor{mygreen}{rgb}{0.0, 0.4, 0.0}

\begin{document}
\makeatletter
\newcommand{\mypm}{\mathbin{\mathpalette\@mypm\relax}}
\newcommand{\@mypm}[2]{\ooalign{%
  \raisebox{.1\height}{$#1+$}\cr
  \smash{\raisebox{-.6\height}{$#1-$}}\cr}}
\makeatother

\twocolumn[
\icmltitle{Defining Benchmarks for Continual Few-Shot Learning}


\begin{icmlauthorlist}
\icmlauthor{Antreas Antoniou}{ed}
\icmlauthor{Massimiliano Patacchiola}{ed}
\icmlauthor{Mateusz Ochal}{ed}
\icmlauthor{Amos Storkey}{ed}
\end{icmlauthorlist}

\icmlaffiliation{ed}{School of Informatics, University of Edinburgh}

\icmlcorrespondingauthor{Antreas Antoniou}{a.antoniou@sms.ed.ac.uk}

\icmlkeywords{Meta Learning, Machine Learning, Deep Learning, Few-shot Learning, Continual Learning, Online Learning, Episodic Learning}

\vskip 0.3in
]

\printAffiliationsAndNotice{}

\begin{abstract}
Both few-shot and continual learning have seen substantial progress in the last years due to the introduction of proper benchmarks. That being said, the field has still to frame a suite of benchmarks for the highly desirable setting of \emph{continual few-shot learning}, where the learner is presented a number of few-shot tasks, one after the other, and then asked to perform well on a validation set stemming from all previously seen tasks. Continual few-shot learning has a small computational footprint and is thus an excellent setting for efficient investigation and experimentation. 
In this paper we first define a theoretical framework for continual few-shot learning, taking into account recent literature, then we propose a range of flexible benchmarks that unify the evaluation criteria and allows exploring the problem from multiple perspectives. As part of the benchmark, we introduce a compact variant of ImageNet, called \emph{SlimageNet64\footnote{Available from  \url{https://zenodo.org/record/3672132}}}, which retains all original 1000 classes but only contains 200 instances of each one (a total of 200K data-points) downscaled to $64 \times 64$ pixels. We provide baselines for the proposed benchmarks using a number of popular few-shot learning algorithms, as a result, exposing previously unknown strengths and weaknesses of those algorithms in continual and data-limited settings.

\end{abstract}
\section{Introduction}
\label{sec:introduction}

Two capabilities vital for an intelligent agent with finite memory are \emph{few-shot learning}, the ability to learn from a handful of data-points, and \emph{continual learning}, the ability to sequentially learn new tasks without forgetting previous ones. There is no question that modern machine learning methods struggle in combining these two capabilities, while humans and animals possess them innately.

\begin{figure}[ht!]
\vspace{-0.2cm}
\centering
\includegraphics[width=1.0\columnwidth,trim={0.8cm 0 0 0}]{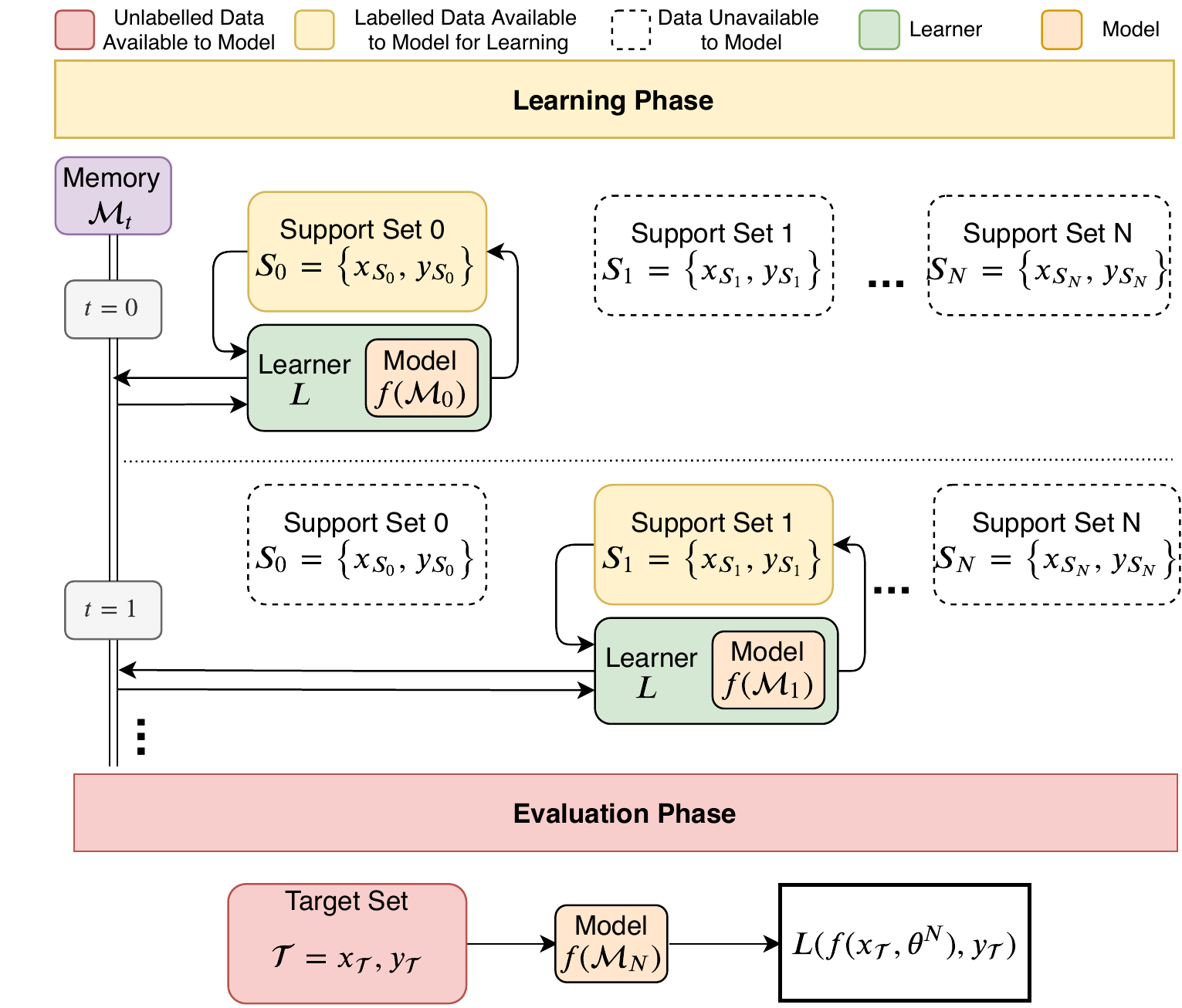}
\caption{High level overview of the proposed benchmark. Top block: from the left, a learner acquires task-specific information from each set, one-by-one, without being allowed to view previous or following sets (memory constraint). The learner can store that knowledge in a shared memory bank. The stored knowledge can be used by a given classification model. On the rightmost side, future tasks are inaccessible to the learner. Central block: the same process is repeated on the second support set. Note that the first support set is now inaccessible. Bottom block: once the learner has viewed all support sets, it is given an evaluation set (target set) containing new examples of classes contained in the support-sets, and tasked with producing predictions for those samples. The evaluation procedure has access to the target set labels and can then establish a generalization measure for the model.}
\vspace{-0.5cm}
\label{fig:overall}
\end{figure}

One of the main reasons behind the scarce consideration of the liaison between the two is that these problems have been often treated separately and handled by two distinct communities. Historically the research on continual learning has focused on the problem of avoiding the loss of previous knowledge when new tasks are presented to the learner, known as \emph{catastrophic forgetting} \cite{mccloskey1989catastrophic}, without paying much attention to the low-data regime. On the other hand, the research on few-shot learning has mainly focused on achieving good generalization over new tasks, without caring about possible future knowledge gain or loss. Scarce attention has been given to few-shot learning in the more practical continual learning scenario.

Taken individually these two areas have recently seen dramatic improvements mainly due to the introduction of proper benchmark tasks and datasets used to systematically compare different methods \cite{chen2019closerfewshot, lesort2019generative, parisi2019continual}. For the set-to-set few-shot learning setting \cite{vinyals2016matching} such benchmarks include Omniglot \cite{lake2015human}, CUB-200 \cite{welinder2010caltech}, Mini-ImageNet \cite{vinyals2016matching} and Tiered-ImageNet \cite{ren2018meta}, whereas for the single-incremental-task continual learning setting \cite{maltoni2019continuous} and the multi-task continual setting \cite{zenke2017continual, lopez2017gradient} the benchmarks include permuted/rotated-MNIST \cite{zenke2017continual,goodfellow2013empirical}, CIFAR10/100 \cite{krizhevsky2009learning}, and CORe50 \cite{lomonaco2017core50}. However, none of those benchmarks are particularly well suited for the constrained task of learning on low-data streams.

Few-shot learning focuses on learning from a small (single) batch of labeled data points. However, it overlooks the possibility of sequential data streams that is inherent in many robotics and embedded systems, as well as standard deep learning training methods, such as minibatch-SGD, where what we have is effectively a sequence of small batches from which a learner must teach an underlying model. On the other hand, continual learning is a broad field encompassing many types of tasks, datasets and algorithms. Continual learning has been applied in the context of general classification \cite{parisi2019continual}, video object recognition \cite{lomonaco2017core50}, and others \cite{lesort2019review}. Most of the investigations are done on continual tasks of very long lengths, using relatively large batches. Moreover, each sub-field has their own combinations of variables (e.g. size and length of sequences) and constraints (e.g. memory, input type) that define groups of continual learning tasks. We argue that our proposed setting helps to formalize and constrain an emerging group of tasks within a low-data setting.

In this paper we propose a setting that bridges the gap between these settings, therefore allowing a spectrum starting from strict few-shot learning going in the middle to short-term continual few-shot learning and on the other end arriving at long-term continual learning. We propose doing this by injecting the sequential component of continual learning into the framework of few-shot learning, calling this new setting \emph{continual few-shot learning}. While we formally define the problem in Section~\ref{sec:cfsl}, a high-level diagram is shown in Figure~\ref{fig:overall}.

In addition to bridging the gap, we argue that the proposed setting can be useful to the research community for four additional reasons. \begin{enumerate*}[label=\bfseries\arabic*.]
    \item As a framework for studying and improving the sample efficiency of mini-batch stochastic methods. Mini-batch training is quite inefficient computationally, because it requires multiple learning iterations over a dataset to learn a good model.
    \item As a minimal and efficient framework for studying and rectifying catastrophic forgetting. Improvements can come in two flavors, either via meta-learning models which can provide insight into better learning dynamics, or by designing general methods to rectify the problem.
    \item As a framework for studying continuous adaptations of neural networks under memory constraints (e.g. robotics, embedded devices)
    \item Due to its continual length and small batch size, CFSL is ideal for investigating and training meta-learning systems that are capable of continual learning. We have made sure that all our settings fit on a single GPU with 11 GBs of memory.
\end{enumerate*}

Our main contributions can be summarized as follows:

\begin{enumerate}[label=\bfseries\arabic*.]
    \item We formalize a highly general and flexible continual few-shot learning setting, taking into account recent considerations and concerns expressed in the literature.
    \item In order to foster a more focused and organized effort in investigating continual few-shot learning, we propose a new benchmark and a compact dataset (SlimageNet64), releasing them under an open source license.
    \item We compare recent state-of-the-art methods on our proposed benchmark, showing how continual few-shot learning is effective in highlighting the strengths and weaknesses of those methods. 
\end{enumerate}

\section{Related Work}
\label{}
\input{related_work.tex}

\begin{figure*}[ht]
\centering
\includegraphics[trim={0 0 0.5cm 0}, width=0.99\linewidth]{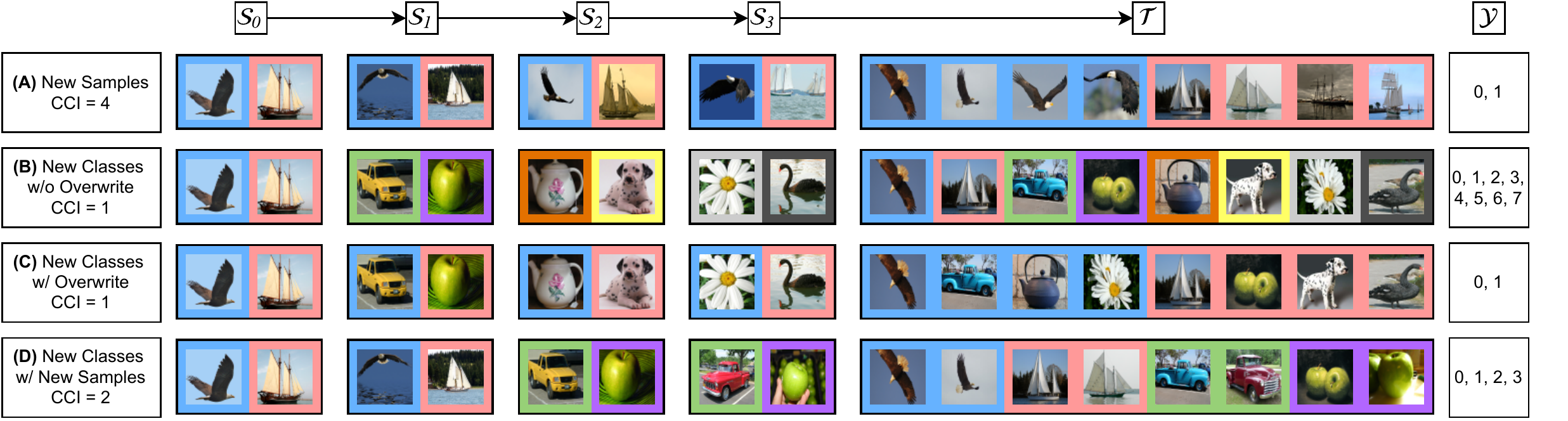}
\caption{Visual representation of the four continual few-shot task types. Each row corresponds to a task with Number of Support Sets, NSS=$4$, and a defined Class-Change Cnterval (CCI). Given a sequence of support sets, $\mathcal{S}_n$, the aim is to correctly classify samples in the target set, $\mathcal{T}$. Colored frames correspond to the associated support set labels.}
\label{fig:task_diagram}
\end{figure*} 

\section{Continual Few-Shot Learning \protect \footnote{A full specification sheet of the proposed setting can be found at \url{https://antreasantoniou.github.io/documents/continual_few_shot_learning_specifiation.pdf}}}\label{sec:cfsl}

This section contains the core contribution of the article. We divide the section in three parts: definition of the problem, where we present a principled formulation of continual few-shot learning; definition of the procedure, where we detail the type of tasks that can be used for learning; definition of the dataset, where we describe the desiderata of a suitable dataset and introduce SlimageNet64.

\subsection{Definition of the problem}
In standard few-shot learning (FSL) for classification a task consists of a small training set (i.e. a support set) $\mathcal{S} = \{(\mathbf{x}_n,y_n)\}_{n=1}^{N_S}$ of input-label pairs, and a small validation set (i.e. a target set) $\mathcal{T} = \{(\mathbf{x}_n,y_n)\}_{n=1}^{N_T}$ of previously unseen pairs. To reduce notation burden we assume that each data-point $x$ has been flattened into a vector of dimensionality $H$. Each label $y \in \mathcal{C}$ with $\mathcal{C}$ being a finite set of classes $\mathcal{C} = \{c_n\}_{n=1}^{N_C} \in \mathbb{N}$. Moreover it is assumed that the pairs in the support and target sets, have different inputs $\mathcal{S}^x \cap \mathcal{T}^x = \{ \emptyset \}$ but same class set $\mathcal{S}^y = \mathcal{C} \land \mathcal{T}^y = \mathcal{C}$, where we have used the shorthand $\mathcal{S}^x = \{\mathbf{x}_n\}_{n=1}^{N_S}$, $\mathcal{S}^y = \{y_n\}_{n=1}^{N_S}$ (likewise for $\mathcal{T}$).
The objective of the learner is to perform well on the validation set $\mathcal{T}$ having only access to the labeled data contained in the support $\mathcal{S}$. The size of the support set $N_S$ is defined by the number of classes $N_C$ (way) and by the \emph{number of samples per class} $K$ (shot), such that if we have a 5-way/1-shot setup we end up with $N_S = N_C \times K = 5 \times 1 = 5$.

In a continual few-shot learning (CFSL) task (i.e. an episode) a single support set is replaced by a sequence of support sets $\mathcal{G} = \{\mathcal{S}_n\}_{n=1}^{N_G}$ with the target set $\mathcal{T} = \{(\mathbf{x}_n,y_n)\}_{n=1}^{N_T}$ now containing previously unseen instances of classes stemming from $\mathcal{G}$. We will refer to $N_G$, the cardinality of $\mathcal{G}$, as the \emph{Number of Support Sets Per Task (NSS)}. Here, each support set in $\mathcal{G}$ contains $N_S$ input-output pairs and is defined as $\mathcal{S} = \{(\mathbf{x}_n,y_n)\}_{n=1}^{N_S}$ like in the standard setup. We also define another parameter, the \emph{Class-Change Interval (CCI)}, that dictates how often the classes should change, in numbers of support sets. This correspond to assign the elements in the support sets to a series of disjoint class sets $\bigcap_{i=1}^{I} \mathcal{C}_{i} = \{ \emptyset \}$. For example, if CCI=2 then we will draw support sets whose classes change every 2 samples. As a result, support sets $\mathcal{S}_1$ and $\mathcal{S}_2$ will contain different instances of the same class set $\mathcal{C}_1$, whereas $\mathcal{S}_3$ and $\mathcal{S}_4$ will contain different instances from the class set $\mathcal{C}_2$. The process of generating CFSL tasks is also described in Algorithm~\ref{algo:cfsl_task} and implemented in the data provider GitHub repository\footnote{\label{url:data-provider-task-generator} The task generator data provider repository can be found at \url{https://github.com/AntreasAntoniou/FewShotContinualLearningDataProvider}}.

A \emph{learner} is a process which extracts task-specific information and distills it into a classification model. The model can be generically defined as a function $f(\mathbf{x}, \boldsymbol{\theta})$ parameterized by a vector of weights $\boldsymbol{\theta}$. At evaluation time the learner is tested through a loss function
\begin{equation}\label{eq_loss}
    \mathcal{L}=\Big( f(\mathbf{x}_{\mathcal{T}}, \boldsymbol{\theta}), y_{\mathcal{T}} \Big),
\end{equation}
where $\mathbf{x}_{\mathcal{T}}$ and $y_{\mathcal{T}}$ are the input-output pairs belonging to the target set. Note that we intentionally provided a definition that is generic enough to fit into different methodologies and not restricted to the use of neural networks.

To remove the possibility of converting a continual learning task to a non-continual one, we introduce a restriction, which dictates that a support set $\mathcal{S}$ is sampled from $\mathcal{G}$ without replacement, and deleted once it has been used by the learner. The learner should never have access to more than one support set at a time, and should not be able to review a support set once it has moved to the next one. This restriction induces a strict sequentiality in the access of $\mathcal{G}$. 

The setup we have described so far is very flexible, and it allows us to define a variety of different tasks and therefore to target different problems. In the following section we provide a description of those tasks and show that they are consistent with the continual learning literature.
\begin{algorithm}[h]
\SetAlgoLined
\KwData{Given labeled dataset $\mathcal{D}$, number of support sets per task $NSS$, number of classes per support set $N_C$, number of samples per support set class $K_S$, number of samples per class for target set $K_T$, class change interval $CCI$, and class overwrite parameter $O$ }
 $a = 1,$\ $b = 1$\;
 
 \For{$a \leq (NSS / CCI)$}{
  Sample and remove $N_C$ classes from $\mathcal{D}$\;
  
  \For{$b \leq CCI$}{
  
  $n \gets a \times CCI + b$
  
  Sample $K_S+K_T$ samples for each of $N_C$ classes
  
  Build support $\mathcal{S}_{n}$ with $K_S$ samples per class\;
  
  Build target $\mathcal{T}_{n}$ with $K_T$ samples per class\;
  
  \uIf{$O = \text{TRUE}$}{
  
    Assign labels $\{1, \dots, N_C\}$ to the classes\;
  
  }
  \Else{
  
  Assign labels $\{1 + (a-1) \times N_C, \dots, N_C \times a\}$ to the classes\;
  
  }
  
  Store sets $\mathcal{S}_{n}$ and $\mathcal{T}_{n}$\;
  }
 }
 
 Combine all target sets $\mathcal{T} = \bigcup_{n=1}^{N_G} \mathcal{T}_{n}$

 Return $(\mathcal{S}_{1...N_G}, \mathcal{T})$\;
 
 \caption{Sampling a Continual Few-Shot Task}\label{algo:cfsl_task}
\end{algorithm}
\subsection{Task Types \protect \footnote{For a full implementation of a task generator data provider see Footnote \ref{url:data-provider-task-generator}}}

In the previous section we have defined the theoretical framework of CFSL, here instead we define an empirical procedure under the form of specific task types. To do so we refer to the literature on continual learning, which has recently focused on more structured procedures, without reinventing the wheel. Note that this is not straightforward, since it is necessary to align the continual learning definitions with the few-shot ones.
In continual learning, there are three generally-accepted scenarios in the context of object recognition \cite{parisi2019continual, lomonaco2017core50}: New Instance (NI), New Class (NC) and New Instance and Class (NIC).

In NI, new patterns of a known set of classes become available with each data batch in a sequence. In NC, new classes become incrementally available. The NIC generalises both types of tasks and incrementally releases patterns of known and new sets of classes. 

Our categorization of CFSL fully covers the standard continual learning setting while introducing an additional, super-class NI setting. Specifically, task \ref{task:NS} and \ref{task:NCw/oO} are equivalent to NI and NC, respectively. Task \ref{task:NCwO} captures the super-set NI setting where instances are sampled across super-classes, instead of being sampled from previously defined class categories. Finally, task \ref{task:NCwNS} explores the NIC setting. Figure \ref{fig:task_diagram} showcases a high-level visual representation of the proposed task.

\input{table_dataset_comparison.tex}

\begin{enumerate}[A]
    
\item{\textbf{New Samples:}}\label{task:NS}

\textbf{Definition:} In this task type, support sets within a given task are sampled from the a preselected set of classes. As a result, any given support set within a task will share the same classes with all other support sets within that task, but will have previously unseen instances (i.e. samples) of those classes:
\begin{equation}
\forall \mathcal{S}_i, \mathcal{S}_j \in \mathcal{G}(
\mathcal{S}_{i}^{x} \cap \mathcal{S}_{j}^{x} = \{ \emptyset \}
\land
\mathcal{S}_{i}^{y} = \mathcal{S}_{j}^{y} = \mathcal{C} ),
\end{equation}
where we have assumed that $\mathcal{S}_{i} \neq \mathcal{S}_{j}$.
For example, if we have 5 classes per support set and 10 support sets, then by the end of the task we have seen 5 classes, each with 10 samples. To achieve this, we can set CCI to be equal to the number of support sets in a given task (CCI = NSS), which means that for every support set we sample new instances and the same classes (as in previous support sets of the same task).

\textbf{Motivation:} Since this setting emulates the default deep learning training regime, it can be useful in studying mini-batch stochastic optimization models as well as meta-learning more efficient algorithms for doing so. It can also be useful when such processes must be executed on a robotic or embedded system.

\item{\textbf{New Classes:}}\label{task:NCw/oO}

\textbf{Definition:} In this task type, each support set has different classes from the other support sets within a given task, formally we write:
\begin{equation}
\forall \mathcal{S}_i, \mathcal{S}_j \in \mathcal{G}(
\mathcal{S}_{i}^{x} \cap \mathcal{S}_{j}^{x} = \{ \emptyset \} 
\land
\mathcal{S}_{i}^{y} \cap \mathcal{S}_{j}^{y} = \{ \emptyset \}),
\end{equation}
with $\mathcal{S}_{i} \neq \mathcal{S}_{j}$.
Similarly to the previous task, here we focus on the case where every class has just a single associate input $\mathbf{x}$ (1-shot). In this task each class within each support set has a corresponding unique output unit in the model. For example, if each support set contains 5 classes and we have 10 support sets, the model will have a total of 50 output units, one for each class. To achieve this, we set CCI to 1, which means that for every task we sample new classes.

\textbf{Motivation:} This setting emulates standard continual learning, where new concepts/classes are acquired as the agent receives a data stream. Therefore it is very useful as a means to investigate such settings or meta-learn models that do well on it. Since this setting allows expanding the number of class descriptors, it is not forced to explicitly rewrite previous knowledge at the class-level, however, it almost always will be required to rewrite representations at lower-levels.

\item{\textbf{New Classes with Overwrite:}}\label{task:NCwO}

\textbf{Definition:} This task is identical to the previous one in terms of how support set inputs are sampled.

The only difference is that the true labels in each support set are \emph{overwritten} by new labels in $C$. This is achieved using the surjective function $O:y\mapsto\widetilde{y}$ that takes as input the labels of a support set $\mathcal{S}^{y}$ and a class set $\widetilde{\mathcal{C}}$, and returns a new support set $\widetilde{\mathcal{S}}=\{(\mathbf{x}_n,\widetilde{y}_n)\}_{n=1}^{N_S}$, with $\widetilde{y} \in \widetilde{\mathcal{C}}$. We can formally write this as: 
\begin{equation}
\begin{split}
\forall \mathcal{S}_i, \mathcal{S}_j \in \mathcal{G}
( \mathcal{S}_{i}^{x} \cap \mathcal{S}_{j}^{x} = \{ \emptyset \} 
\land
\mathcal{S}_{i}^{y} \cap \mathcal{S}_{j}^{y} = \{ \emptyset \}
\land \\
O(\mathcal{S}_{i}^{y}, \widetilde{C}) = O(\mathcal{S}_{j}^{y}, \widetilde{C}) = \widetilde{\mathcal{S}}_{i}^{y} = \widetilde{\mathcal{S}}_{j}^{y} = \widetilde{C}),
\end{split}
\end{equation}
where $\mathcal{S}_{i} \neq \mathcal{S}_{j}$.
This task is similar to task~\ref{task:NS} in terms of the number of output units, however, in task~\ref{task:NCwO} a single output unit is associated with more than one true class. Intuitively, $\widetilde{\mathcal{C}}$ could be the hierarchical categories of classes in $\mathcal{G}^y = \cup_{n=1}^{N_G} \mathcal{S}^{y}_{n}$, however, we assign the hierarchical categories arbitrarily.

In practical terms, if we have 5 classes and 10 support sets, in this task the model only uses 5 output units to store all 50 classes. Therefore, for every support set the output unit of a specific class is overwritten with a new one. To obtain this task we need to set CCI to 1, then apply the overwrite function.

\textbf{Motivation:} This setting emulates situations where an agent is tasked with learning data-streams while being limited in storing that knowledge in a preset number of output classification labels. As a result the agent learns super classes. This setting is useful in investigating how effective a learner is in continually updating a class descriptor while not forgetting previous descriptions. Since this setting does not allow expanding the number of class descriptors, it is forced to explicitly rewrite previous knowledge at the class-level, with which certain types of models might struggle more than others. This setting is especially useful for robotics and embedded system applications.

\item{\textbf{New Classes with New Samples:}}\label{task:NCwNS}

\textbf{Definition:} In this task type, the sampled support sets contain different instances of the same classes for some predefined CCI ($1 <$ CCI $<$ NSS) such that
\begin{equation}
\forall \mathcal{S}_i, \mathcal{S}_j \in \mathcal{G}(\mathcal{S}_{i}^{y} = \mathcal{S}_{j}^{y} \leftrightarrow \mathcal{S}_i \in \mathcal{G}_k \land \mathcal{S}_j \in \mathcal{G}_k),
\end{equation}
where $\mathcal{G}_k$ is a partition of the task set $\mathcal{G}$ satisfying

\begin{equation}
|\mathcal{G}_k| = \text{CCI}, \
\bigcap_{k=1}^{N_G/\text{CCI}} \mathcal{G}_k = \{ \emptyset \}, \
\bigcup_{k=1}^{N_G/\text{CCI}} \mathcal{G}_k = \mathcal{G}.
\end{equation}
Note that this partitioning ensures that the subsets are pairwise disjoint.
If we have 5 classes per support set, 10 support sets and a CCI of 5, we end up with 5 support sets containing samples from 5 classes and other 5 support sets containing samples from 5 different classes. This makes a total of 10 classes, each one containing 5 samples.

\textbf{Motivation:} This setting emulates situations where an agent is tasked with both learning new class descriptors and updating such descriptors by observing new class instances. This setting sheds light on how agents can perform on a setting that mixes all previous settings into one.

\end{enumerate}

\subsection{Metrics}

In this section we provide a number of metrics useful in comparing different models applied to the CSFL setting. It is important to note that each one of this metrics only provides a quantifier for a desirable property. Whether a model is superior to another can only be said when comparing them on the same metric. Whether a model is more desirable than another depends on the task and hardware that a system is trying to solve. 

\subsubsection{Test Generalization Performance}
A proposed model should be evaluated on at least the test sets of Omniglot and SlimageNet, on all the tasks of interest. This is done by presenting the model with a number of previously unseen continual tasks sampled from these test sets, and then using the target set metrics as the task-level generalization metrics. To obtain a measure of generalization across the whole test set the model should be evaluated on a number of previously unseen and unique tasks. The mean and standard deviation of both accuracy and performance should be used as generalization measures to compare models.

\subsubsection{Across Task Memory (ATM)}
Even though we have imposed a restriction on the access to $\mathcal{G}$, the learner is still authorized to store in a local \emph{memory bank} $\mathcal{M}$ some representations of the inputs and/or output vectors (often implemented as embedding vectors or inner loop parameters)
\begin{equation}\label{eq_memory}
    \mathcal{M}=\{ (\hat{\mathbf{x}}, \hat{y})_{\mathcal{S}_1}, ..., (\hat{\mathbf{x}}, \hat{y})_{\mathcal{S}_{N_G}} \},
\end{equation}
where $\hat{\mathbf{x}}$ and $\hat{y}$ are representations of $\mathbf{x}$ and $y$ obtained after a given learner has processed $\mathbf{x}$ and $y$ and stored some of their useful components. Most learners will be compressing a given support set, but this is not strictly the case.

Note that the potential compression rate is not directly correlated to the complexity of the model (e.g. number of parameters, FLOPs, etc). For instance, compression can be achieved by removing some of the dimensions of the input, or by using a lossless data compression algorithm, which may not require additional parameters or may have minimal impact on the execution time. In this regard, the concept of memory bank $\mathcal{M}$ helps to disambiguate model complexity from any additional memory allocated for compressed representations of inputs. We can use the cardinality of $\mathcal{M}$, indicated as $|\mathcal{M}|$, to quantify the learner efficiency. Given two learners with their corresponding models $f(\mathbf{x}, \boldsymbol{\theta}_1)$ and $f(\mathbf{x}, \boldsymbol{\theta}_2)$, and assuming that the size of $\boldsymbol{\theta}_1$ is equal to the size of $\boldsymbol{\theta}_2$ with $\mathcal{L}_1 = \mathcal{L}_2$, then the learner with smaller cardinality $|\mathcal{M}|$ must be preferred.

In order to compare performances across different tasks and datasets, we relate the size of the stored task-specific representations (in bytes) $\mathcal{M}^{\hat{x}}$ (e.g. embedding vectors in ProtoNets, and inner loop parameters for MAML) during task-specific information extraction to the size of vectors (in bytes) $\mathbf{x}$ contained in the episode $\mathcal{G}^x = \cup_{n=1}^{N_G} \mathcal{S}^{x}_{n}$. Recall that $\hat{\mathbf{x}}$ is a compressed version of $\mathbf{x}$ and therefore $F < H$. To reduce the notation burden we have only considered the inputs $\mathbf{x}$ and not the targets $y$, since $\mathbf{x}$ is significantly larger than $y$. Based on these considerations we define Across-Task Memory (ATM)
\begin{equation}\label{eq_task_memory_ratio}
    \text{ATM} = \frac{|\mathcal{M}^{\hat{x}}|}{|\mathcal{G}^x|},
\end{equation}
where $\mathcal{M}^{\hat{x}}$ is the stored representation of a series of support sets and $\mathcal{G}^x$ is the size of the support sets. Note that for each utilized floating point arithmetic unit we include a computation that takes into account the floating point precision level. For example, if both  $\mathcal{M}^{\hat{x}}$ and $\mathcal{G}^x$ use the same floating point standard then it is divided out, but if the representational form uses a lower precision than the actual data-points then it becomes compressive.
From a practical standpoint (image classification), the ATM can be estimated relating the total number of bytes stored in the memory bank (ATM numerator) with the total number of bytes over all the images in the episode (ATM denominator). Given the definition above: $\text{ATM}<1$ for learners with efficient memory, $\text{ATM}=0$ for learners with no memory, and $\text{ATM}>1$ for learners with inefficient memory. Note that the ATM is undefined for empty episodes~$\mathcal{G} = \{ \emptyset \}$. ATM is task/dataset agnostic and can be used to compare various models (or the same model) across different settings.

To summarize ATM is useful for the following reasons:

\begin{enumerate}
    \item We do not restrict our agents to a specific amount of memory for their continual task learning. As a result, an agent could easily store whole support sets into its memory bank. We want to be able to distinguish between more memory efficient models (that might in compress support sets efficiently) and less memory efficient models.
    \item Using default measures of computational capacity such as MACs is not enough. MACs do not quantify the actual memory shared across the learning process, but instead quantifies the overall computational requirements of the models. Such memory requirements might be minuscule when compared to the model architecture functions which are usually orders of magnitude more expensive. Therefore there is a need for a quantifier that focuses on the efficiency of the learner at compressing incoming data, and how that varies with additional number of support sets.
\end{enumerate}

\subsubsection{Multiply-Addition operations (MACs)}
This  metric measures the computational expense of both the learner and the model operations during learning and inference time. This is different than ATM, as ATM reflects how much memory is required to store information about a support when the next support set is observed, whereas the inference memory footprint measures the memory footprint that the model itself needs to execute during one cycle of inference, and meta-learning cycle.

\subsubsection{FSL vs CFSL vs CL} 

At this point, it is important to properly explain what the relationship between FSL, CSFL and CL is. We argue that all three belong in a spectrum within which the free variables are size of an incoming support set, and the number of support sets within a task. If the size of a support set is very small, e.g. five samples consisting of a single sample from five classes and the number of support sets is one, then we have few-shot learning. If we increase the number of support sets to more than one up to a hundred steps, we have CFSL. Once we begin to increase the size of the support set to something reminiscent of standard deep learning training (e.g. within the range of 32-256 where most models are trained) and we increase the number of support sets into the thousands, we end up with the full continual learning setting. 

\input{results_table.tex}
\subsection{Datasets}\label{sec:datasets}
Properly training and evaluating a CFSL agent can be an arduous process. Building such tasks requires datasets that meet the following desiderata:

\begin{enumerate}
    \item \textbf{Diversity:} An optimal dataset should have a very high degree of diversity in terms of classes. This enforces robustness in the learning procedure, since the model has to be able to deal with previously unseen class semantics. In addition, diversity enable the training, validation, and test splits to lie within different distribution spaces, covering classes that are significantly different from one another. 
    \item \textbf{Number of classes:} The dataset should contain a very high number of categories. This is to ensure that we can train models on CFSL tasks ranging from 1 sub-task, all the way to 100s of sub-tasks. \textcolor{black}{Ideally, the length of a sub-task sequence should not be constrained by the number of classes in the dataset.}
    \item \textbf{Number of samples per class:}
    \textcolor{black}{The dataset should contain a fair, but not overabundant, number of samples per class. On the one hand, a dataset with few samples can not capture the difference in distribution within each class, resulting in a poor evaluation measure. Moreover, training a learner on a small dataset can produce significant underfitting issues. On the other hand, having too many samples per class increases the training time, producing very strong learners but neutralizing the difference among them. As a result, it would be much harder to draw any conclusion on the capabilities of the underlying algorithms, since the difference in performance between them would be minimal.}
    \item \textbf{Size:} \textcolor{black}{Finally, we would like our models to be trained in reasonable time, finances and computational resources. Thus, the size of the dataset should be contained, such that it can be easily managed and stored in memory. This requirement is crucial to allow use of the dataset by a significant portion of the research community. Here, we define a dataset as appropriate if its size does not exceed 16 GB, which is our reasonable estimate of the average laptop RAM.}
\end{enumerate}

Many datasets already exist in continual and few-shot learning, however most of them do not satisfy all the aforementioned requisites and are insufficient for robust benchmarking of CFSL algorithms. Omniglot \cite{lake2015human} was a good first choice for a lower-difficulty dataset, however, we were still missing a higher complexity dataset with coloured images.

For this reason we propose a new variant of ImageNet64$\times$64 \cite{chrabaszczimagenet17}, named  \emph{SlimageNet64} (derived from Slim and ImageNet). SlimageNet64 consists of 200 instances from each of the 1000 object categories of the ILSVRC-2012 dataset \cite{krizhevsky2012imagenet, russakovsky2015ilsvrc}, for a total of 200K RGB images with a resolution of $64 \times 64 \times 3$ pixels. We created this dataset from the downscaled version of ILSVRC-2012, ImageNet64x64, as reported in \cite{chrabaszczimagenet17}, using the \emph{box} downsampling method available from \emph{Pillow} library.
In Table~\ref{table:dataset_comparison} we report a detailed comparison of all the datasets available, showing how SlimageNet64 is an optimal choice in terms of diversity, number of classes, number of samples per class, and storage size. The closest alternative to SlimageNet64 is Tiered-ImageNet \cite{ren2018meta}, a subset of ILSVRC-12 with a total of 608 classes. Comparing the two, SlimageNet64 contains more classes and overall has a higher class diversity across train, validation, and test. Moreover, it has a lower computational footprint due to the smaller resolution of the images and the lower number of samples per class. These characteristics make SlimageNet64 more compact and at the same time more challenging.

\begin{figure*}
    \centering
    \includegraphics[width=1.0\linewidth]{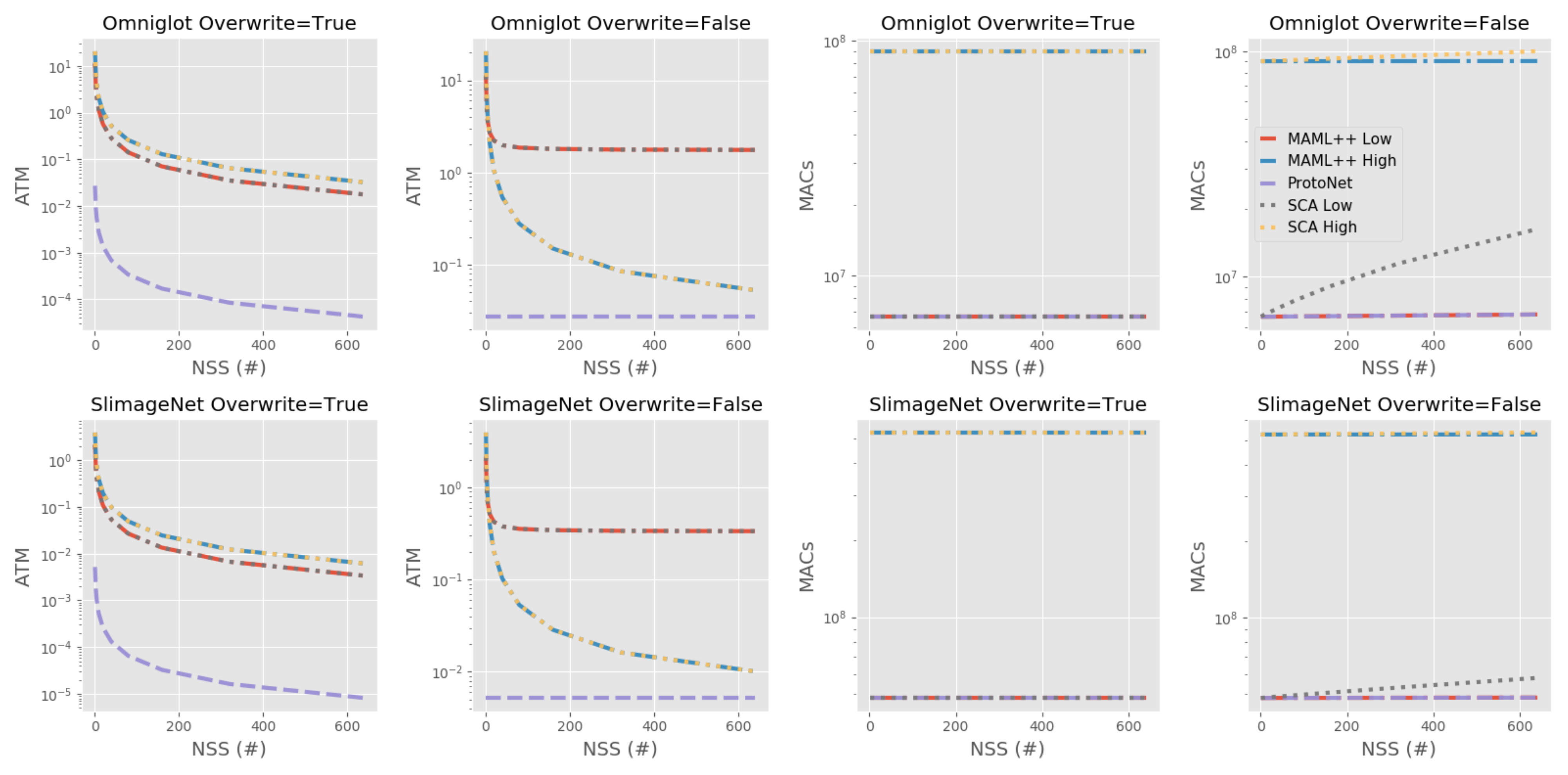}
    \caption{ATM (Across-Task Memory) and MAC (Multiply-Accumulate Computations) costs for a variety of NSS (Number of Support Sets Per Task). ProtoNets are the superior method across the board. In terms of ATM it is worth noting that methods such as MAML++ H and SCA tend to become incrementally cheaper than MAML++ L as the number of support sets increases. Whereas in terms of MACs MAML++ H and SCA are the most expensive by an order of magnitude or more compared to MAML++ L and ProtoNets.}
    \label{fig:atm_mac_cost}
\end{figure*}

\section{Experiments \protect \footnote{We provide an implemetation that reproduces all the experiments in this section at \url{https://github.com/AntreasAntoniou/FewShotContinualLearning}}}

For the purposes of establishing baselines in the CFSL tasks outlined in this paper we chose to use six existing FSL methods: (i) randomly initializing a convolutional neural network, and fine tuning on incoming tasks, (ii) pretraining a convolutional neural network on all training set classes and then fine-tune on sequential tasks \cite{chen2019closerfewshot}, (iii) Prototypical Networks \cite{snell2017prototypical} (baseline for metric-based FSL methods), (iv) the Improved Model Agnostic Meta-Learning or MAML++ L (Low-End) model \cite{antoniou2018train} (baseline for optimization based FSL methods), (v) MAML++ H (High-End) model \cite{antoniou2019learning} (dense-net backbone, squeeze excite attention, mid-tier baseline), and (vi) the Self-Critique and Adapt model (SCA) \cite{antoniou2019learning}, a top state-of-the-art algorithm for FSL (high-tier baseline). For each model, we used the exact configurations specified in their original papers. For each method (apart from ProtoNets) we used five inner-loop update steps. 

\begin{figure*}[t]
\centering
\includegraphics[width=1.0\linewidth]{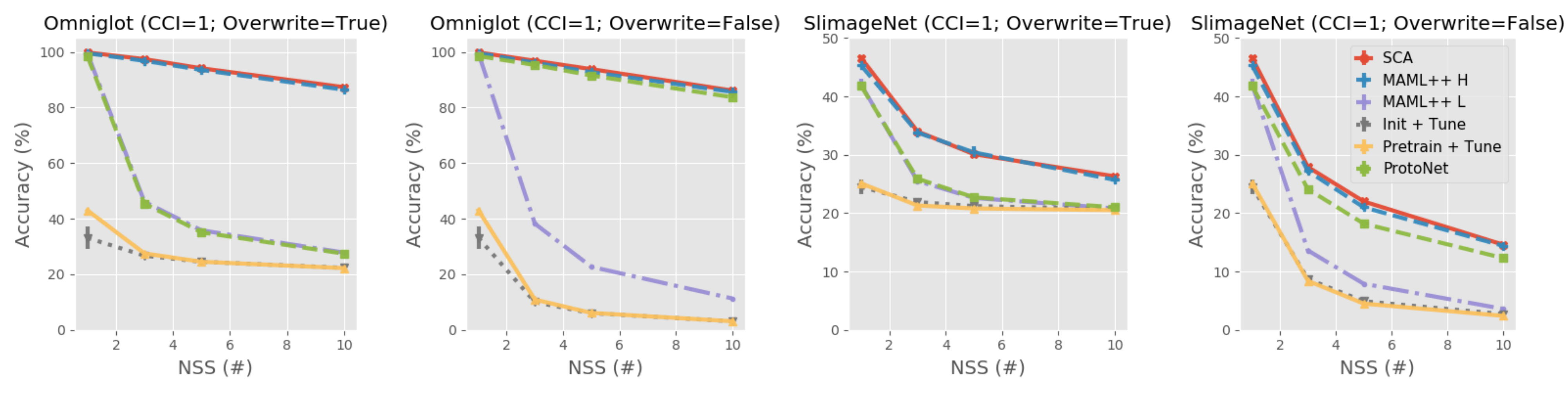} 
\caption{Accuracy (percentage) of different methods on the Omniglot and SlimageNet datasets for different values of Number of Support Sets Per Task (NSS). We report both with/without overwrite. This figure illustrates which methods tend to be more robust to increasing NSS (SCA, MAML ++ H) and which methods do not (ProtoNets, MAML++ L, Init/Pretrain + Tune), as well as to how sensitive they are to those changes.} \label{fig:results}
\end{figure*} 

For each continual learning task type, we ran experiments on each dataset. Each support set contained 1 sample from 5 classes (5-way, 1-shot) while the target sets contained 5 samples from all the classes seen in a given task. We ran experiments using 1, 3, 5 and 10 support sets for each continual task, therefore creating tasks of increasingly long number of sub-tasks. We ran each experiment 3 times, each time with different seeds for the data-provider and the model initializer. All models were trained for 250 epochs, where each epoch consisted of 500 update steps, each one done on a single continual task, using the default configuration of the Adam learning rule, and weight-decay of 1e-5. At the end of each training epoch we validated a given model by applying it on 600 randomly sampled continual tasks, keeping those tasks consistent across all validation phases. Once all epochs have been completed, we built an ensemble of the top five models across all epochs with respect to validation accuracy, and applied that on 600 random tasks sampled from the test set, to compute the final performance metrics.

For Omniglot, we used the first 1200 classes for the training set, and we split the rest equally to create a validation and test set. For SlimageNet64, we used 700, 100 and 200 classes to build our training, validation and test sets respectively. The SlimageNet64 splits were chosen such that the training set had mostly living organisms, with some additional everyday tools and buildings, while the validation and test sets contained largely inanimate objects. This was done to ensure sufficient domain-shift between the training and evaluation distributions. As a result this enables a more robust generalization measure to be computed. 

\section{Baseline Results}
Results are reported in Table~\ref{table:results} and Figure~\ref{fig:results}. The results from our proposed benchmark, have revealed previously unknown weaknesses and strengths of existing few-shot learning methods. In Omniglot, in the New Classes without Overwrite Setting (\ref{task:NCw/oO}) MAML++ Low-End is inferior to ProtoNets, whilst in the New Classes with Overwrite Settings (\ref{task:NCwO}) this result is reversed. From this we can infer that embedding-based methods are better at retaining information from previously seen classes, assuming that each new class remains distinct. However, when overwriting is enabled this trend is overturned because ProtoNet prototypes are shared by a number of super-classes containing classes that are harder to semantically disentangle. Gradient based methods such as MAML++ dominate in this setting, since they can update their weights towards new tasks, and therefore achieve a better disentanglement of those super-classes. SCA and High-End MAML++ (which utilize both embeddings and gradient-based optimization) produce the best performance across all settings. In the New Samples Setting (\ref{task:NS}), gradient based methods tend to outperform embedding-based methods while hybrid methods produce the best results. Furthermore, in the New Classes and Samples Setting (\ref{task:NCwNS}), embedding-based methods outperform gradient-based methods, whilst hybrid methods continue to produce the best performing models.

In SlimageNet, ProtoNets seem to consistently outperform the Low-End MAML++ model, even in the New Classes with Overwrite Settings (\ref{task:NCwO}) where it was previously inferior. This might indicate that in SlimageNet retaining information about previously seen tasks is more important than disentangling complicated super-classes. Overall models that use both embedding-based and gradient-based methods, seem to outperform methods that do just one of the two often with a performance boost of 100-200\%. In the New Classes and Samples Setting (\ref{task:NCwNS}), embedding-based methods outperform gradient-based ones by a significant margin, while hybrid approaches consistently generate the best performing models. Interestingly, in the New Samples Setting (\ref{task:NS}) using SlimageNet64, the embedding-based and gradient-based methods produce very similar results to one another, whereas in Omniglot gradient-based methods dominate.

Furthermore Figure \ref{fig:atm_mac_cost} shows the ATM and MAC costs for a range of NSS, starting from one, up to and including 640. Some notable observations include the fact that ProtoNets are simply the most efficient in both metrics, by two orders magnitude. In addition, even though the Low-End MAML++ starts off cheaper than the high end model, as NSS increases, it eventually becomes far more expensive than the High-End variant. This is mostly due to the fact that the Low-End MAML++ flattens its features and applies a linear layer at the output side of the network. As a result, for each additional new class to be learned, there is one magnitude higher cost than the high-end MAML which simply global pools its features before applying a linear layer.

\section{Conclusion}
In this paper, we have introduced a new flexible and extensive benchmark for Continual Few-shot Learning. We have also introduced a new minimal variant of ImageNet, called SlimageNet64, that contains all of ImageNet classes, but only 200 samples from each class, downscaled to 64$\times$64. The dataset requires just 9 GB of RAM, and it can be easily loaded in memory for faster experimentation. Furthermore, we have run experiments on the proposed benchmarks, utilizing a number of popular few-shot learning models and baselines. In doing so, we have found that embedding-based models tend to perform better when incoming tasks contain different classes from one another, potentially due to better task-specific information retention. On the other hand, gradient-based methods tend to perform better when the task-classes form super-classes of randomly combined classes, resulting in a disentangled task that is harder to predict. Gradient-based methods work better here thanks to their ability of dynamic adaptation, whereas more static methods like ProtoNets tend to produce poorer performances. That being said, in datasets of higher class diversity and sample complexity, gradient-based methods perform like embedding-based methods. We assume that this is due to the nature of the data, making class-information retention more relevant than disentanglement factors. Methods utilizing both embedding-based and gradient-based methods (i.e. High-End MAML++ and SCA) outperform methods that use either of the two. In conclusion, we hope that the proposed benchmark and dataset, will help increasing the rate of progress and the understanding of the behavior of systems trained in a continual and data-limited setting.

\section{Acknowledgements}
We would like to thank Elliot Crowley, Paul Micaelli, Eleanor Platt, Ondrej Bohdal, Sen Wang, and Joseph Mellor for reviewing this work and providing useful suggestions/comments. This work was supported in part by the EPSRC Centre for Doctoral Training in Data Science, funded by the UK Engineering and Physical Sciences Research Council (Grant No. EP/L016427/1) and the University of Edinburgh as well as a Huawei DDMPLab Innovation Research Grant. Furthermore, additional funding for the project was provided by a joint grant by the UK Engineering and Physical Sciences Research Council and SeeByte Ltd (Grant No. EP/S515061/1).

\bibliography{metalearning,deeplearning,datasets,continuallearning}
\bibliographystyle{icml2020}

\end{document}

%% file: related_work.tex
\subsection{Few-Shot Learning}
Progress in few-shot learning (FSL) was greatly accelerated after the introduction of the set-to-set few-shot learning setting \cite{vinyals2016matching}. This setting, for the first time, formalized few-shot learning as a well defined problem paving the way to the use of end-to-end differentiable algorithms that could be trained, tested, and compared. What followed was an explosion of progress in the field. Among the first algorithms to be proposed there were meta-learned solutions, which here we group into three categories: 

\textbf{Embedding-learning and Metric-learning:} Those methods include the Neural Statistician \cite{edwards2016towards}, Matching Networks \cite{vinyals2016matching} and Protoypical Networks \cite{snell2017prototypical}. They are based on the idea of parameterizing embeddings via neural networks and then use distance metrics to match target points to support points in latent space. The whole process is fully differentiable and it is trained such that the model can generalize to a wide range of tasks.

\textbf{Optimization-based or Gradient-based Meta-Learning:} Those methods have been introduced in the form of MetaLearner LSTM \cite{ravi2016optimization}, MAML  \cite{finn2017model}, Meta-SGD \cite{li2017meta} and MAML++ \cite{antoniou2018train}. The model itself is a model for learning, explicitly trained to achieve a particular set of tasks. More specifically, in such models there is an inner-loop optimization process that is partially or fully parameterized with fully differentiable modules. This inner-loop process is optimized such that if a model uses it to learn from a support set, then it will generalize to a target set. The process that learns the learner is the outer-loop optimization process. This mechanism of learning to learn, is often called \emph{meta-learning} \cite{schmidhuber1987evolutionary}. Recent methods such as LEO \cite{rusu2018meta} and SCA \cite{antoniou2019learning} have combined both categories to create very strong state-of-the-art systems.

\textcolor{black}{\textbf{Hallucination-based:} Those algorithms can utilize one or both the aforementioned methods in combination with a generative process, to produce additional samples as a complement to the support set. An example of this approach has been recently presented by \citet{antoniou2017data}.}

\textcolor{black}{\textbf{Other solutions:} There have been a number of methods that do not clearly fall in one of the previous categories. One example are Bayesian approaches, like those based on amortized networks \cite{gordon2019meta}, hierarchical models \cite{grant2018recasting}, or Gaussian Processes \cite{patacchiola2019deep}.
Another example are Relational Networks \cite{santoro2017simple}, originally created to deal with relational reasoning; they have been adapted to the few-shot learning setting with good performance \cite{santurkar2018does}. In addition, simpler approaches such as pretraining of a neural network on all classes and fine tuning on a given support set, have also shown to perform fairly well \cite{chen2019closerfewshot}. Similarly, a method based on nearest neighbor classifier has recently showed to achieve state-of-the-art performances \cite{wang2019simpleshot}.}

\subsection{Continual Learning}
The problem of continual learning (CL), also called life-long learning, has been considered since the beginnings of artificial intelligence and it remains an open challenge in robotics \cite{lesort2019review} and machine learning \cite{parisi2019continual}. In standard supervised learning, algorithms can access any data point as many times as necessary during the training phase. In contrast, in CL data arrives sequentially and can only be provided once during the training process. Following the taxonomy of \citet{maltoni2019continuous} we group the continual learning methods into three classes: architectural, rehearsal, and regularization methods.

\textbf{Architectural methods:} Architectural strategies add, clone, or save parts of trained weights \cite{lesort2019generative}. For example, progressive neural networks \cite{rusu2016progressive} create a new neural network for each new task and connect it to previously generated networks, thus leveraging previously learned knowledge while solving catastrophic forgetting. Another architectural strategy includes weight freezing \cite{mallya2018piggyback, mallya2018packnet} where some weights are frozen dynamically to retain knowledge of old tasks, while leaving others to freely adapt to new tasks later on.

\textbf{Rehearsal methods:} Rehearsal strategy methods selectively choose which data points to store within a bounded amount of resources. One such algorithm stores top-$N$ most representative samples of a class while maintaining a fixed upper bound on the required memory \cite{rebuffi2017icarl}. More recently, generative models such as GANs and VAEs \cite{lesort2018state, lesort2019marginal} have been proposed to represent previously seen data as weights of a neural network.

\textbf{Regularization methods:} Unlike other approaches, regularization methods focus on adding constraints on parameter updates of neural networks to directly minimize catastrophic forgetting. For example, Elastic Weight Consolidation (EWC, \citealt{kirkpatrick2017overcoming, mitchell2018never}) slows down the learning rate of those weights that are responsible for previously learned tasks. Other regularization techniques have been recently presented which follow a similar approach \cite{zenke2017continual, lee2017toward, he2018overcoming}. 

All of the outlined approaches offer various advantages and disadvantages under resource constraints. Architectural approaches can be constrained on the amount of available RAM, whereas, rehearsal strategies can become quickly bounded by the amount of available storage. Regularization approaches can be free from resource constraints but incur in severe issues in the way they adapt model parameters. Note that the outlined strategies are not mutually exclusive and can be combined \cite{rebuffi2017icarl, maltoni2019continuous, kemker2018measuring}.

\emph{Online learning} is a special case of CL where new data becomes available a single data point at a time. \emph{Active learning} can also appear in continual learning settings but it is a special type of semi-supervised machine learning, that aims to strategically select unlabeled data points for future labeling in order to maximize accuracy while reducing the amount of input provided by the user.


\subsection{Inconsistencies in the evaluation protocol}

In the literature does not exist a proper benchmark that integrates few-shot and continual learning. Some related tasks were hastily introduced as a mean to prove the efficacy of a given system, making such tasks very restricted in terms of what methods they are applicable on and how many aspects they can investigate. We found that tasks and datasets vary from paper to paper, making it challenging to know the actual performance of a given algorithm. For instance, the method proposed by \citet{vuorio2018meta}, an extention of MAML able to act as a loss function in the inner loop of the algorithm, has been tested exclusively on variants of MNIST. The method of \citet{javed2019meta}, an online meta-objective that minimize catastrophic forgetting, has been tested on Omniglot and incremental sine-waves. The work of \citet{finn2019online}, another extension of MAML to the online setting, has been evaluated on MNIST, CIFAR-100 and PASCAL 3D+. These inconsistencies in the evaluation protocol of continual few-shot algorithms further support our proposal of a unified benchmark.

Related to continual few-shot learning is the field of \emph{incremental few-shot learning} \cite{qiao2018few, gidaris2018dynamic}. The difference between the two lies in how the target sets are sampled during the evaluation phase. In incremental few-shot learning the end performance of trained models is evaluated on target sets sampled from classes encountered at meta-training phase as well as new classes sampled from the evaluation dataset. In continual few-shot learning, during evaluation, support and target sets are sampled only from the test set. Incremental and continual few-shot learning are tangential, the two share similar objectives but are significantly different in terms of training and testing procedures. For this reason we will not analyze this line of research any further.

In conclusion, from this literature analysis it is evident how the problem of continual few-shot learning is not well defined, making it challenging to benchmark and compare performance of algorithms. In the next section, we will focus on formalizing the problem and then we will propose a unified set of tasks and datasets to encourage consistent benchmarking.

%% file: table_dataset_comparison.tex
\begin{table*}
\caption{Dataset comparisons. Dataset details include: number of classes in the whole dataset (\textbf{\# Classes}), number of samples per class (\textbf{\# Samples}), total number of images (\textbf{\# Total}), \textbf{Resolution}, \textbf{Format}, and finally, \textbf{Size} indicating the allocation of RAM for the whole dataset. Suitability include: class diversity (\textbf{Diversity}), enough classes (\textbf{\# Classes}), enough samples (\textbf{\# Samples}), proper size (\textbf{Size}). Omniglot and SlimageNet64 are the best choices for the tasks on grayscale and RGB datasets, respecitively, according to our suitability criteria (for details see section~\ref{sec:datasets}). }
\centering
\scalebox{0.7}{
\begin{tabular}{l|cccccc|cccc} \hline
 & \multicolumn{6}{c|}{\textbf{Dataset details}} & \multicolumn{4}{c}{\textbf{Suitability (satisfies criteria)} } \\
 \textbf{Dataset} & \textbf{\# Classes} & \textbf{\# Samples} & \textbf{\# Total} & \textbf{Resolution} & \textbf{Format} & \textbf{Size (GB)} & \textbf{Diversity} & \textbf{\# Classes} & \textbf{\# Samples} & \textbf{Size} \\ \hline
MNIST  \cite{lecun1998mnist} & 10 & 7000 & 70k & 28$\times$28 & Grayscale & $\sim$0.20 & \xmark & \xmark & \xmark & \cmark \\
Fashion MNIST  \cite{xiao2017fashion} & 10 & 7000 & 70k & 28$\times$28 & Grayscale & $\sim$0.20 & \xmark & \xmark & \xmark & \cmark \\
\textbf{Omniglot}  \cite{lake2015human} & 1622 & 20 & $\sim$32.4k & 28$\times$28 & Grayscale & $\sim$0.095  & \cmark & \cmark & \cmark & \cmark \\ \hline
CUB-200  \cite{welinder2010caltech} & 200 & 20-39 & 6033 & $\sim$475$\times\sim$400 & RGB & $\sim$13 & \xmark & \xmark & \xmark & \cmark \\
Mini-ImageNet  \cite{vinyals2016matching} & 100 & 600 & 60k & 84$\times$84 & RGB & $\sim$4.7 & \xmark & \xmark & \cmark & \cmark \\
Tiered-ImageNet  \cite{ren2018meta} & 608 & 600 & $\sim$365k & 84$\times$84 & RGB & $\sim$29 & \cmark & \cmark & \cmark & \xmark \\
CIFAR-100  \cite{krizhevsky2009learning} & 100 & 600 & 60k & 32$\times$32 & RGB & $\sim$0.68 & \xmark & \xmark  & \cmark & \cmark \\
CORe50  \cite{lomonaco2017core50} & 10 & $\sim$16.5k & $\sim$165k & 128$\times$128 & RGB-D & $\sim$30 & \xmark & \xmark & \xmark & \xmark \\
ILSVRC2012  \cite{russakovsky2015ilsvrc} & 1000 & 732-1300 & $\sim$1.43M & 224$\times$224 & RGB & $\sim$800 & \cmark & \cmark &  \xmark & \xmark \\
\textbf{SlimageNet64  (ours)} & 1000 & 200 & 200k & 64$\times$64 & RGB & $\sim$9.1 & \cmark & \cmark & \cmark & \cmark  \\ \hline
\end{tabular}
}
\label{table:dataset_comparison}
\end{table*}


%% file: results_table.tex
\begin{table*}[ht!]
\caption{Accuracy and standard deviation (percentage) on the test set for the proposed benchmarks and tasks. Best results in bold.}
\label{table:results}
\centering

\scalebox{0.61}{
    \begin{tabular}{cl|cccccccccccc}
    \hline
    & Task Type & FSL & \ref{task:NCw/oO} & \ref{task:NCwO} & \ref{task:NS} & \ref{task:NCwNS} & \ref{task:NCw/oO} & \ref{task:NCwO} & \ref{task:NS} & \ref{task:NCwNS} & \ref{task:NCw/oO} & \ref{task:NCwO} & \ref{task:NS} \\
    & NSS & 1 & 3 & 3 & 3 & 4 & 5 & 5 & 5 & 8 & 10 & 10 & 10 \\
    & CCI & 1 & 1 & 1 & 3 & 2 & 1 & 1 & 5 & 2 & 1 & 1 & 10 \\
    & Overwrite & - & False & True & True & False & False & True & True & False & False & True & True \\  \hline    
    
    \parbox[t]{1mm}{\multirow{6}{*}{\rotatebox[origin=c]{90}{\textbf{Omniglot}}}} &
    Init + Tune & $43.05$\scalebox{0.75}{$ \mypm0.01$}&    $10.87$\scalebox{0.75}{$ \mypm0.01$}&	$27.51$\scalebox{0.75}{$ \mypm0.01$}&	$44.76$\scalebox{0.75}{$ \mypm0.01$}&	$8.74$\scalebox{0.75}{$ \mypm0.01$}&	$~6.15$\scalebox{0.75}{$ \mypm0.01$}&	$24.52$\scalebox{0.75}{$ \mypm0.01$}&	$45.30$\scalebox{0.75}{$ \mypm0.01$}&	$~3.93$\scalebox{0.75}{$ \mypm0.01$}&	$~3.12$\scalebox{0.75}{$ \mypm0.01$}&	$22.16$\scalebox{0.75}{$ \mypm0.01$}&	$45.64$\scalebox{0.75}{$ \mypm0.01$}	\\
    
    & Pretrain + Tune & $33.07$\scalebox{0.75}{$ \mypm2.04$}&	$~9.97$\scalebox{0.75}{$ \mypm0.14$}&	$26.75$\scalebox{0.75}{$ \mypm0.27$}&	$32.44$\scalebox{0.75}{$ \mypm0.29$}&	$~7.91$\scalebox{0.75}{$ \mypm0.15$}&	$~6.02$\scalebox{0.75}{$ \mypm0.02$}&	$24.51$\scalebox{0.75}{$ \mypm0.06$}&	$31.89$\scalebox{0.75}{$ \mypm1.10$}&	$~3.86$\scalebox{0.75}{$ \mypm0.06$}&	$~3.13$\scalebox{0.75}{$ \mypm0.03$}&	$22.30$\scalebox{0.75}{$ \mypm0.06$}&	$33.17$\scalebox{0.75}{$ \mypm0.39$}	\\
    
    & ProtoNets  & $98.52$\scalebox{0.75}{$ \mypm0.04$}&    $95.30$\scalebox{0.75}{$ \mypm0.12$}&	$45.44$\scalebox{0.75}{$ \mypm0.19$}&	$98.73$\scalebox{0.75}{$ \mypm0.02$}&	$48.98$\scalebox{0.75}{$ \mypm0.03$}&	$91.52$\scalebox{0.75}{$ \mypm0.20$}&	$35.10$\scalebox{0.75}{$ \mypm0.09$}&	$98.73$\scalebox{0.75}{$ \mypm0.12$}&	$48.44$\scalebox{0.75}{$ \mypm0.03$}&	$83.72$\scalebox{0.75}{$ \mypm0.19$}&	$27.39$\scalebox{0.75}{$ \mypm0.17$}&	$98.65$\scalebox{0.75}{$ \mypm0.14$} \\
    
    & MAML++ L  & $99.46$\scalebox{0.75}{$ \mypm0.03$}&    $38.18$\scalebox{0.75}{$ \mypm0.14$}&	$46.12$\scalebox{0.75}{$ \mypm0.15$}&	$99.38$\scalebox{0.75}{$ \mypm0.07$}&	$28.87$\scalebox{0.75}{$ \mypm0.07$}&	$22.69$\scalebox{0.75}{$ \mypm0.07$}&	$35.76$\scalebox{0.75}{$ \mypm0.14$}&	$99.41$\scalebox{0.75}{$ \mypm0.04$}&	$14.29$\scalebox{0.75}{$ \mypm0.05$}&	$11.30$\scalebox{0.75}{$ \mypm0.02$}&	$27.82$\scalebox{0.75}{$ \mypm0.03$}&	$99.44$\scalebox{0.75}{$ \mypm0.01$}	\\
    
    & MAML++ H & $99.54$\scalebox{0.75}{$ \mypm0.03$}& $96.14$\scalebox{0.75}{$ \mypm0.02$}&	$96.77$\scalebox{0.75}{$ \mypm0.08$}&	$99.73$\scalebox{0.75}{$ \mypm0.04$}&	$49.44$\scalebox{0.75}{$ \mypm0.02$}&	$92.70$\scalebox{0.75}{$ \mypm0.03$}&	$93.47$\scalebox{0.75}{$ \mypm0.05$}&	$99.80$\scalebox{0.75}{$ \mypm0.01$}&	$49.00$\scalebox{0.75}{$ \mypm0.04$}&	$85.56$\scalebox{0.75}{$ \mypm0.10$}&	$86.38$\scalebox{0.75}{$ \mypm0.14$}&	$99.86$\scalebox{0.75}{$ \mypm0.01$}	\\
    
    & SCA  & $\mathbf{99.78}$\scalebox{0.75}{$ \mathbf{\mypm0.01}$}&	$\mathbf{96.84}$\scalebox{0.75}{$ \mathbf{\mypm0.04}$}&	$\mathbf{97.38}$\scalebox{0.75}{$\mathbf{ \mypm0.02}$}&	$\mathbf{99.82}$\scalebox{0.75}{$ \mathbf{\mypm0.01}$}&	$\mathbf{49.71}$\scalebox{0.75}{$ \mathbf{\mypm0.01}$}&	$\mathbf{93.81}$\scalebox{0.75}{$ \mathbf{\mypm0.02}$}&	$\mathbf{94.08}$\scalebox{0.75}{$ \mathbf{\mypm0.45}$}&	$\mathbf{99.88}$\scalebox{0.75}{$ \mathbf{\mypm0.03}$}&	$\mathbf{49.51}$\scalebox{0.75}{$ \mypm0.01$}&	$\mathbf{86.07}$\scalebox{0.75}{$ \mathbf{\mypm0.03}$}&	$\mathbf{87.29}$\scalebox{0.75}{$ \mathbf{\mypm0.19}$}&	$\mathbf{99.88}$\scalebox{0.75}{$\mathbf{ \mypm0.01}$} \\\hline

    \parbox[t]{1mm}{\multirow{6}{*}{\rotatebox[origin=c]{90}{\textbf{SlimageNet64}}}} &
    Init + Tune & $25.1$\scalebox{0.75}{$ \mypm0.01$}&	$~8.4$\scalebox{0.75}{$ \mypm0.01$}&	$21.3$\scalebox{0.75}{$ \mypm0.01$}&	$24.4$\scalebox{0.75}{$ \mypm0.01$}&	$~6.1$\scalebox{0.75}{$ \mypm0.01$}&	$~4.5$\scalebox{0.75}{$ \mypm0.01$}&	$20.8$\scalebox{0.75}{$ \mypm0.01$}&	$24.7$\scalebox{0.75}{$ \mypm0.01$}&	$~3.0$\scalebox{0.75}{$ \mypm0.01$}&	$~2.4$\scalebox{0.75}{$ \mypm0.01$}&	$20.5$\scalebox{0.75}{$ \mypm0.01$}&	$24.9$\scalebox{0.75}{$ \mypm0.01$}	\\
    
    & Pretrain + Tune & $24.5$\scalebox{0.75}{$ \mypm0.60$}&	$~8.7$\scalebox{0.75}{$ \mypm0.03$}&	$21.9$\scalebox{0.75}{$ \mypm0.11$}&	$24.2$\scalebox{0.75}{$ \mypm0.17$}&	$~6.4$\scalebox{0.75}{$ \mypm0.01$}&	$~4.9$\scalebox{0.75}{$ \mypm0.02$}&	$21.2$\scalebox{0.75}{$ \mypm0.05$}&	$24.5$\scalebox{0.75}{$ \mypm0.23$}&	$~3.3$\scalebox{0.75}{$ \mypm0.03$}&	$~2.7$\scalebox{0.75}{$ \mypm0.03$}&	$20.7$\scalebox{0.75}{$ \mypm0.10$}&	$24.4$\scalebox{0.75}{$ \mypm0.20$}	\\

    & ProtoNets & $41.8$\scalebox{0.75}{$ \mypm0.16$}&	$24.1$\scalebox{0.75}{$ \mypm0.05$}&	$25.9$\scalebox{0.75}{$ \mypm0.23$}&	$43.1$\scalebox{0.75}{$ \mypm0.24$}&	$15.1$\scalebox{0.75}{$ \mypm0.03$}&	$18.2$\scalebox{0.75}{$ \mypm0.14$}&	$22.7$\scalebox{0.75}{$ \mypm0.09$}&	$43.3$\scalebox{0.75}{$ \mypm0.03$}&	$10.4$\scalebox{0.75}{$ \mypm0.12$}&	$12.3$\scalebox{0.75}{$ \mypm0.09$}&	$21.0$\scalebox{0.75}{$ \mypm0.06$}&	$43.7$\scalebox{0.75}{$ \mypm0.15$} \\

    & MAML++ L & $42.0$\scalebox{0.75}{$ \mypm0.48$}&	$13.6$\scalebox{0.75}{$ \mypm0.04$}&	$25.5$\scalebox{0.75}{$ \mypm0.23$}&	$42.7$\scalebox{0.75}{$ \mypm0.10$}&	$10.2$\scalebox{0.75}{$ \mypm0.11$}&	$~7.9$\scalebox{0.75}{$ \mypm0.13$}&	$22.6$\scalebox{0.75}{$ \mypm0.03$}&	$43.0$\scalebox{0.75}{$ \mypm0.12$}&	$~5.0$\scalebox{0.75}{$ \mypm0.08$}&	$~3.6$\scalebox{0.75}{$ \mypm0.14$}&	$20.8$\scalebox{0.75}{$ \mypm0.09$}&	$43.0$\scalebox{0.75}{$ \mypm0.42$}	\\
    
    & MAML++ H & $45.3$\scalebox{0.75}{$ \mypm0.14$}&	$27.2$\scalebox{0.75}{$ \mypm0.25$}&	$33.8$\scalebox{0.75}{$ \mypm0.16$}&	$61.2$\scalebox{0.75}{$ \mypm0.36$}&	$16.8$\scalebox{0.75}{$ \mypm0.18$}&	$21.0$\scalebox{0.75}{$ \mypm0.21$}&	$\mathbf{30.4}$\scalebox{0.75}{$ \mathbf{\mypm0.51}$}&	$68.6$\scalebox{0.75}{$ \mypm0.47$}&	$12.3$\scalebox{0.75}{$ \mypm0.11$}&	$14.4$\scalebox{0.75}{$ \mypm0.12$}&	$25.7$\scalebox{0.75}{$ \mypm0.10$}&	$75.6$\scalebox{0.75}{$ \mypm0.10$}	\\

    & SCA & $\mathbf{46.6}$\scalebox{0.75}{$\mathbf{ \mypm0.16}$}&	$\mathbf{27.9}$\scalebox{0.75}{$ \mathbf{\mypm0.16}$}&	$\mathbf{34.0}$\scalebox{0.75}{$\mathbf{ \mypm0.23}$}&	$\mathbf{65.3}$\scalebox{0.75}{$\mathbf{ \mypm0.15}$}&	$\mathbf{17.3}$\scalebox{0.75}{$\mathbf{ \mypm0.07}$}&	$\mathbf{22.0}$\scalebox{0.75}{$\mathbf{ \mypm0.18}$}&	$30.1$\scalebox{0.75}{$\mypm0.36$}&	$\mathbf{72.0}$\scalebox{0.75}{$\mathbf{ \mypm0.36}$}&	$\mathbf{12.7}$\scalebox{0.75}{$\mathbf{ \mypm0.08}$}&	$\mathbf{14.6}$\scalebox{0.75}{$\mathbf{ \mypm0.07}$}&	$\mathbf{26.3}$\scalebox{0.75}{$\mathbf{ \mypm0.13}$}&	$\mathbf{77.4}$\scalebox{0.75}{$\mathbf{ \mypm0.06}$} \\ \hline
    
    \end{tabular}
    }

\end{table*}